\pdfoutput=1
\documentclass[fleqn,10pt,twocolumn]{SICE13}
\usepackage{graphicx}        
\usepackage{multicol}        
\usepackage[bottom]{footmisc}
\usepackage{url}

\usepackage{times}

\usepackage{subfigure} 
\usepackage{amssymb}
\usepackage[cmex10]{amsmath}

\usepackage[a4paper,top=72pt, bottom=72pt, left=57pt, right=57pt]{geometry}

\title{Contingency Training}

\author{Danilo Vasconcellos Vargas${}^{1}$, Hirotaka Takano${}^{2}$ and Junichi Murata${}^{3}$}

\affils{${}^{1}$Graduate School of Information Science and Electrical Engineering, Kyushu University, Fukuoka, Japan\\
(E-mail: vargas@cig.ees.kyushu-u.ac.jp)\\
${}^{2}$Faculty of Information Science and Electrical Engineering, Kyushu University, Fukuoka, Japan\\
(E-mail: hirotaka@cig.ees.kyushu-u.ac.jp)\\
${}^{3}$Faculty of Information Science and Electrical Engineering, Kyushu University, Fukuoka, Japan\\
(E-mail: murata@cig.ees.kyushu-u.ac.jp)\\
}
\abstract{ 
When applied to high-dimensional datasets, feature selection algorithms might still leave dozens of irrelevant variables in the dataset.
Therefore, even after feature selection has been applied, classifiers must be prepared to the presence of irrelevant variables.
This paper investigates a new training method called Contingency Training which increases the accuracy as well as the robustness against irrelevant attributes.
Contingency training is classifier independent.
By subsampling and removing information from each sample, it creates a set of constraints.
These constraints aid the method to automatically find proper importance weights of the dataset's features.
Experiments are conducted with the contingency training applied to neural networks over traditional datasets as well as datasets with additional irrelevant variables.
For all of the tests, contingency training surpassed the unmodified training on datasets with irrelevant variables and even outperformed slightly when only a few or no irrelevant variables were present.
}

\keywords{Irrelevant Variables, Contingency Training, Classification, Neural Networks, Feature Weighting, Dimensional Reduction.}

\begin{document}

\maketitle


\section{Introduction}
\label{introduction}

Real world classification problems often involve multiple features (the feature space is high dimensional). 
Beyond being computationally expensive, these problems possess the curse of dimensionality which make them harder for various reasons (distance metrics become less useful, sampling becomes more expensive as the volume grows exponentially with the number of dimensions, etc...) \cite{aggarwal2005k,friedman1997bias,indyk1998approximate}.
One way of alleviating the curse of dimensionality is to use feature selection algorithms to decrease the number of features 
\cite{biesiada2005feature}.

Feature selection methods has been applied for some time and their success is widely known 
\cite{yang1997comparative,guyon2003introduction,weston2000feature}.
There are other advantages of employing feature selection, to cite some: speeding up the learning process, facilitating the understanding of the model. 
However, it is difficult to select exactly all the relevant features. 
To avoid the risk of excluding relevant variables and losing information, dozens of irrelevant variables might still remain in the dataset.


To prepare a classifier to the eventual presence of irrelevant variables, in this paper, we propose a new training method for supervised learning hereby called Contingency Training.
The purpose of contingency training is to improve the accuracy and robustness to irrelevant variables of any classifier by modifying its training dataset. 
The proposed method creates from the initial dataset a more difficult and bigger dataset with some of its values missing.
Actually, the proposed method generates a mixture of constraints with the interesting property that they do not possess a bias, since they are generated by a uniform distribution.

The following is highlighted as the most salient advantages of the approach:
\begin{itemize}
\item Robustness against irrelevant variables;
\item Simplicity - The contingency training can be implemented easily by just modifying the samples from the training dataset;
\item Accuracy - Provides a relevant increase in accuracy;
\item Classifier independent - It works for any classifier algorithm without any required modification in the classifier itself.
\end{itemize}

Contingency training is not a feature selection algorithm.
Actually, both methods have different objectives and should work together to enable a system to solve high dimensional problems.
In one hand, the objective of feature selection is to remove irrelevant features and decrease dimensionality, on the other hand, contingency training aims to improve the performance of any classifier on the presence of irrelevant variables.

\section{Related Methods}

Functionally, contingency training relates to the idea of feature weighting.
The idea of weighting the importance of variables is not new. 
It was developed in various methods for feature selection \cite{guyon2003introduction,gnanadesikan1995weighting}. 
However, few are the methods which can use a feature's importance weighting vector directly (such as the k-means \cite{huang2005automated}).
Most of the methods can only make use of a weighting vector by setting a threshold and removing the least important variables \cite{weston2001feature,blum1997selection}.
Contigency training is the first to perturbate the dataset to force feature weighting inside the learning phase, which is algorithm independent and can make any supervised learning algorithm take the importance of the variables into account.

Structurally, the proposed method relates to methods that pre-process the dataset \cite{kotsiantis2007supervised}.
But there is not any dataset pre-processing method with similar objective to the proposed method. 
Their objectives range from efficiency improvement for large datasets (e.g., subsampling the dataset to decrease its overall size \cite{kniss2001interactive}) to privacy preservation (e.g., perturbating the dataset to preserve the privacy while preserving important information \cite{chen2005privacy}). 

\section{Contingency Training}
\label{gen_inst}


Let $S$ be the final set of training samples to train a given classifier. 
Here, $n$ and $nv$ are defined to be respectively the number of samples and the number of variables.
$\{s_1, ..., s_n\}$ are the initial training samples where $s_i= \{v_1^i, ...,v_{nv}^i\}$ defines a sample composed of $nv$ variables ($v$). 
Using these definitions, the contingency training is explicitly described in Table~\ref{training_alg}.
\begin{table}[h]
 \centering
\caption{Contingency Training Algorithm} 
\begin{tabular}{p{7cm}}
\hline
\begin{enumerate}
\item $S = \{s_1, ..., s_n\}$
\item $CS = \emptyset$
\item while(checkCriterion())
\begin{enumerate}
\item $newsample = replaceWithMissing(S)$
\item $CS= CS\cup~\{newsample\}$
\end{enumerate}
\item $S=CS\cup~S$
\end{enumerate} \\
\hline
\end{tabular}
\label{training_alg}
\end{table}

Where the function for creating missing values ($replaceWithMissing()$) replaces some values in the dataset with missing values uniformly with probability $prob$ over all variables. 
And the criterion to stop repeating samples ($checkCriterion()$) may be set to a maximum constant over the size of the dataset or even based on the learning error of the algorithm. 
In this article, we will use a maximum constant for the $checkCriterion()$. 
The next Section will cover a notation for specifying the contingency training which determines precisely both the maximum constant and the probability $prob$. 

\subsection{Notation}
\label{notation}

This Section describes the notation which specifies exactly how the dataset is created in the tests. 
Let $p_a$ be the size of the artificial dataset (composed of only artificial samples) created from samples of the initial dataset by {\em replaceWithMissing()}, and consider $p_i$ the original size of the initial dataset.
Yet, let $r_a=\frac{p'_a}{p_i}$ and $r_i=\frac{p'_i}{p_i}$, where $p'_a$ and $p'_i$ are respectively the size of the artificial and initial datasets that were used to compose the final dataset $S$. Note that usually $p'_a=p_a$, but $p'_i=0$ may often hold true and therefore $p'_i \ne p_i$.
Now it is possible to define a notation $r_aA/r_iI/prob$ defining uniquely a given contingency training setting, where $prob$ is the same probability of replacing values by missing values explained in Section~\ref{gen_inst}.


\subsection{Missing Value Representation}

\begin{figure}[h]
 \centering
 \includegraphics[scale=0.4]{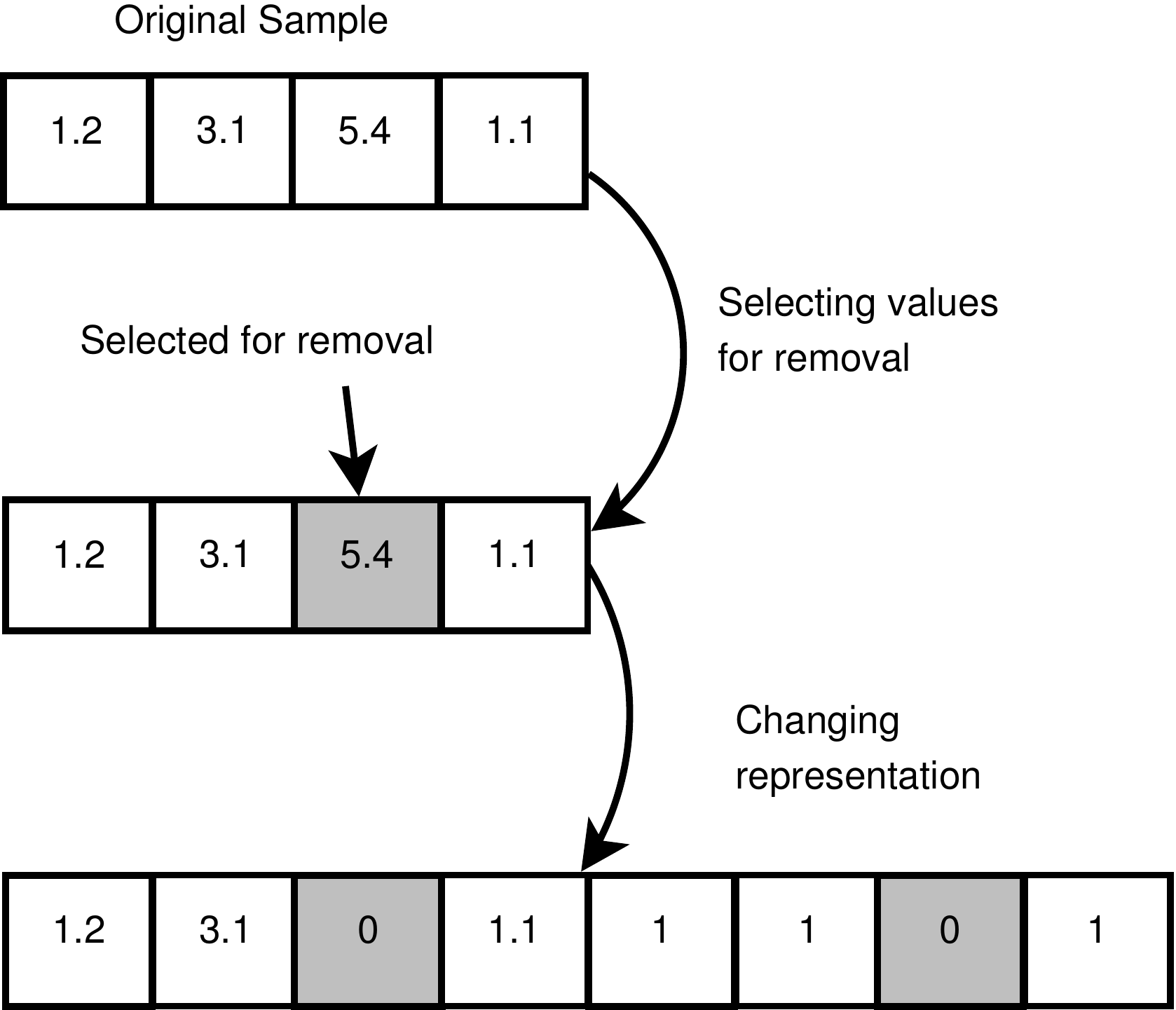}
 \caption{Example of how the method replace non missing values with missing values and construct the final representation.}
 \label{missing_representation}
\end{figure}

In the sections above, it was pointed out that some values are replaced by missing values, but no specific information was given on how to represent the missing values explicitly.
This paper represents every missing value by the zero value independent of the dataset. 
Additionally, it concatenates each sample with a binary vector with the same size of the sample itself, containing respectively for each variable of the original sample either one (if the respective value is not missing) or zero (if the respective value is missing).
Figure~\ref{missing_representation} gives an example of the representation used.

This representation is used to provide the dataset with enough information to enable the learning even when datasets have originally plenty of zero values. In preliminary tests we observed that the concatenation performed better than other forms of representation (such as adding an offset value).





\section{Experiments}


\subsection{Settings}

The tests are conducted with $100$ trials using different $75\%/25\%$ train/test splits (training dataset has a size of 75\% of the entire dataset while the testing dataset have the remaining 25\%) over the following datasets from the UCI machine learning repository \cite{uci2010}: Glass \footnote{The dataset from UCI has an index as the first attribute and an ordered class. This index was removed to avoid a trivial direct mapping from index to class.}, Wine, Zoo, Iris and ``Pima Indians Diabetes" (which we will refer to as ``Diabetes") and an additional dataset called Az-5000 corresponding to a character recognition problem created by \cite{az-dataset}.
The number of classes, samples and variables of each dataset are provided in Table~\ref{datasets}.
Moreover, in Table~\ref{datasets} there are two datasets (Iris and Diabetes) which were modified to contain $20$ additional irrelevant variables created from a uniform distribution $Unif(0,\alpha)$, where $\alpha$ is chosen randomly from another uniform distribution $Unif(1,20)$ and fixed for all samples of the same input.

\begin{table*}
\centering
\caption{Characteristics of the Datasets}
\resizebox{12cm}{!}
{
\begin{tabular}{|c|c|c|c|}
\hline
	Datasets & Number of variables & Number of classes & Number of samples \\ \hline
	Wine & 13 & 3 & 178 \\ \hline 
	Zoo & 16 & 7 & 102 \\ \hline
	Diabetes & 8 & 2 & 768 \\ \hline
	Glass & 10 & 7 & 214 \\ \hline 
	Az-5000 & 18 & 26 & 5000 \\ \hline 
	Modified Iris & 24 & 3 & 150 \\ \hline
	Modified Diabetes & 28 & 2 & 768 \\ \hline
\end{tabular}
}
\label{datasets}
\end{table*}

For the classifier implementation we used the nnet package from R \cite{nnet} which is a feed-forward multilayer perceptron (MLP) \cite{haykin2009neural,rumelhart1995backpropagation} having a single hidden layer and Broyden-Fletcher-Goldfarb-Shanno (BFGS) as learning algorithm \cite{shanno1985broyden}.
Table~\ref{parameters} shows the parameters for the neural network. 
The contingency training parameters will be defined per test using the notation defined in Section~\ref{notation}.

\begin{table}
\centering
\caption{Neural Network's Parameters}
\begin{tabular}{|c|c|}
\hline
	Initial weights & $[-0.1,0.1]$ \\ \hline 
	Weight decay & $1e-4$ \\ \hline
	Hidden nodes & 15 \\ \hline
	Maximum iterations & 1000000 \\ \hline 
	Output units & logistic function \\ \hline 
\end{tabular}
\label{parameters}
\end{table}

\subsection{Tests over Datasets with Irrelevant Variables}

This section focuses on datasets with irrelevant variables.
Figure~\ref{challenge_complex} shows the results for the two datasets with additional irrelevant variables (Iris and Diabetes) as well as the Az-5000 dataset (character recognition task). 
Exceptionally, the test on the Az-5000 dataset had ten trials instead of the usual $100$ trials.

With the proposed method the median increased of $6.5\%$, $5.2\%$, $5.2\%$, $7.7\%$, $5.5\%$ and $7.6\%$ respectively for the Az-5000, Iris, Diabetes, Wine, Glass and Zoo datasets. 
These results demonstrate that contingency training is a promising way of preparing classifiers to the eventual presence of irrelevant variables.
The reason of why the proposed method surpasses the usual training method will be explained in Section~\ref{explanation}.

\begin{figure*}[!htb]
 \centering
 \includegraphics[scale=0.55]{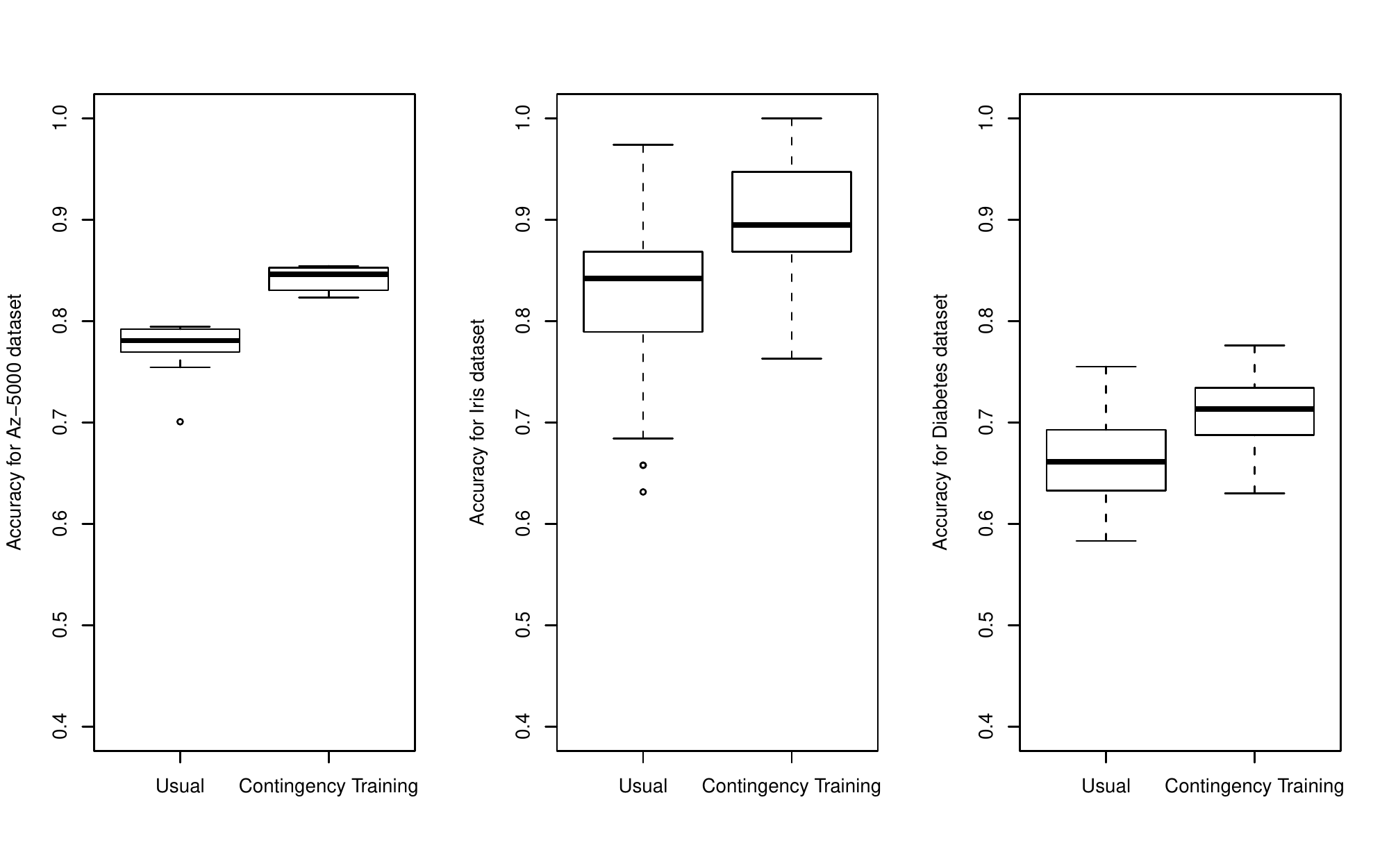}
 \caption{Accuracy on (from left to right) Az-5000, Iris and Diabetes (both Iris and Diabetes have $20$ additional irrelevant variables) datasets using an unmodified dataset (usual training) and contingency training with a $10A/0I/0.1$ setting.}
 \label{challenge_complex}
\end{figure*}

\begin{figure*}[!htb]
 \centering
 \includegraphics[scale=0.55]{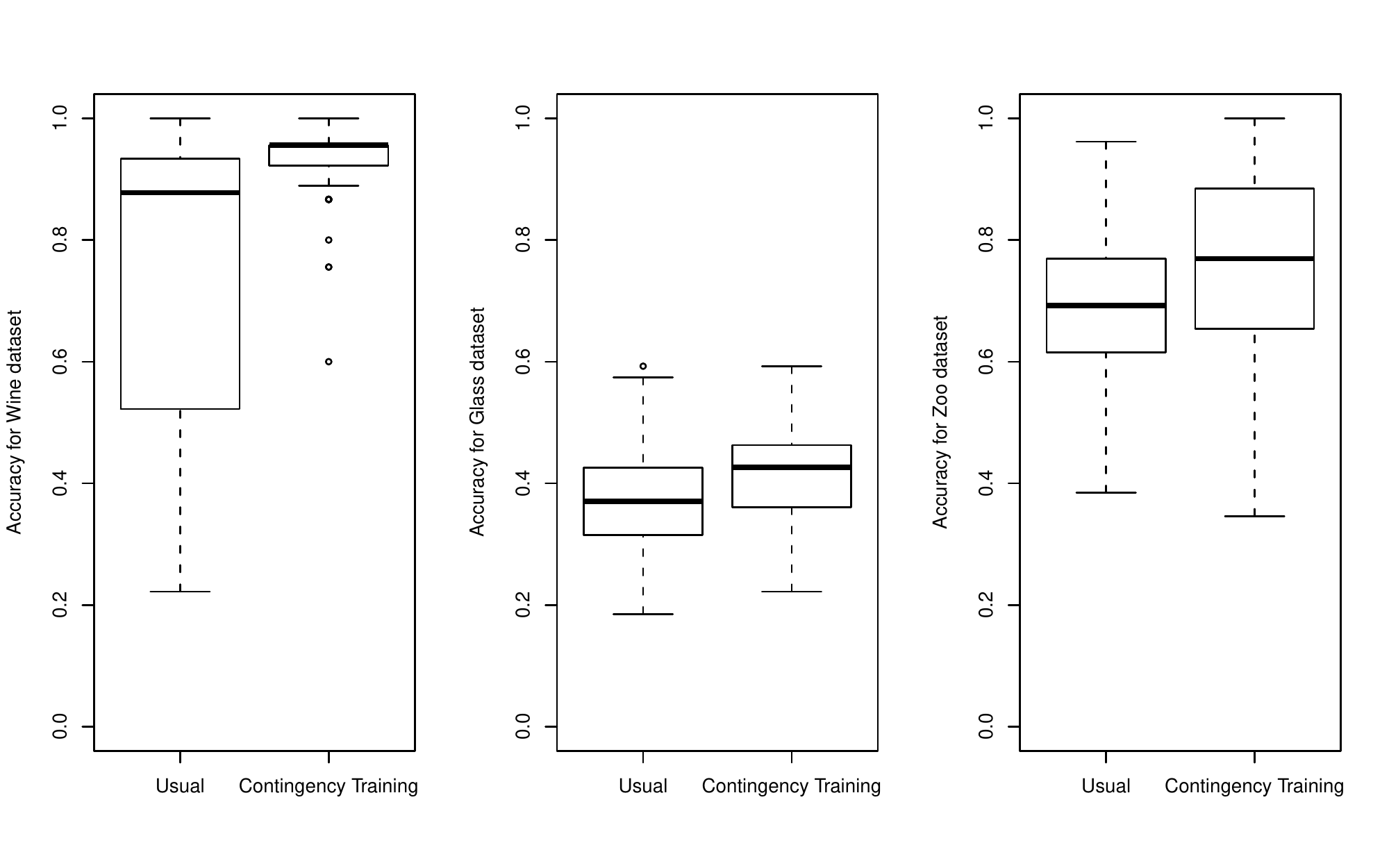}
 \caption{Accuracy on (from left to right) Wine, Glass and Zoo (all of them possess $20$ additional irrelevant variables) datasets using an unmodified dataset (usual training) and contingency training with a $10A/0I/0.1$ setting.}
 \label{challenge_complex}
\end{figure*}

\subsection{Tests over Traditional Datasets}

\begin{figure*}[!htb]
 \centering
 \includegraphics[scale=0.55]{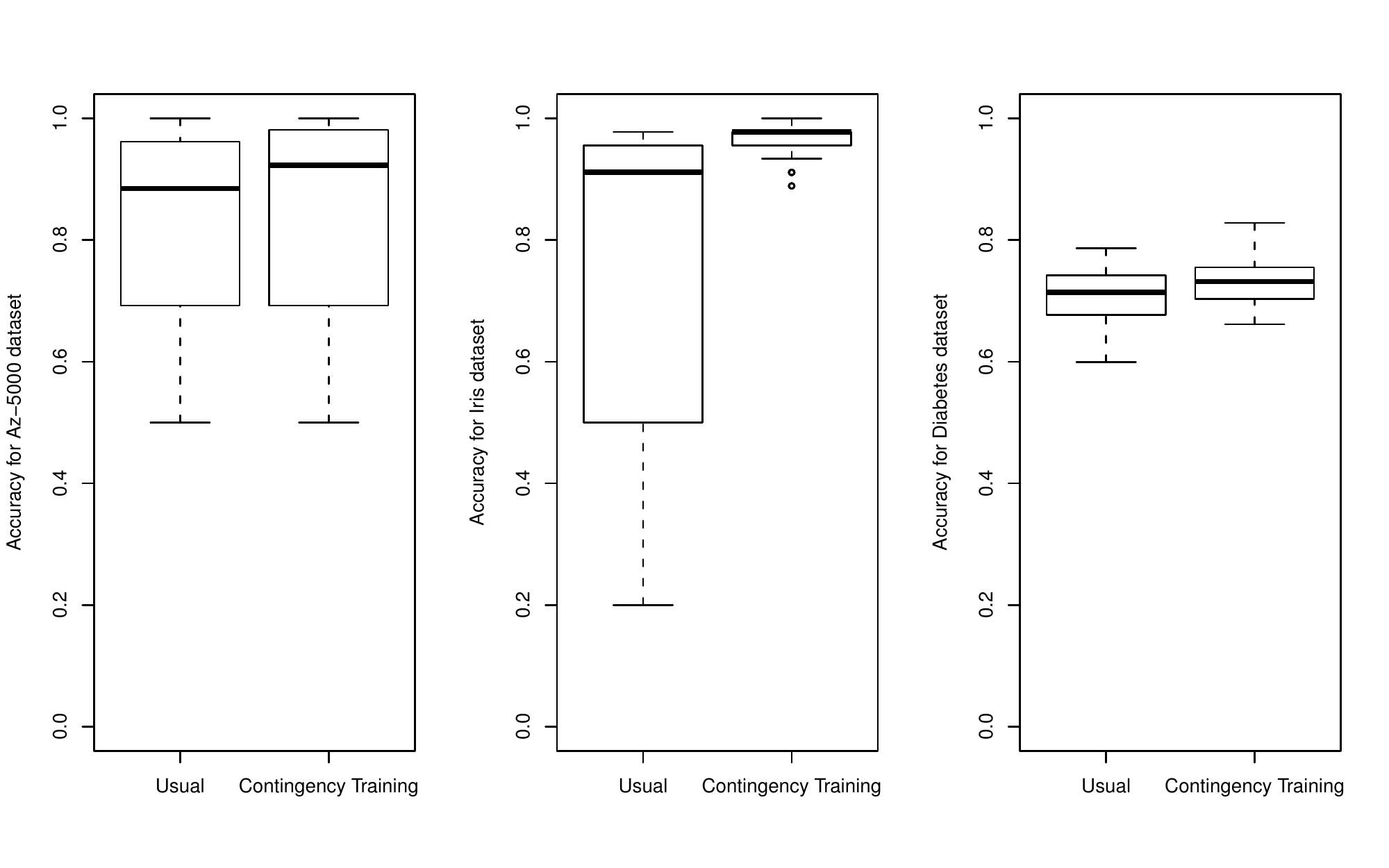}
 \caption{Accuracy on (from left to right) Zoo, Wine and Diabetes datasets using an unmodified dataset (usual training) and contingency training with a $1A/1I/0.1$ setting.}
 \label{challenge_1_1}
\end{figure*}

\begin{figure*}[!htb]
 \centering
 \includegraphics[scale=0.55]{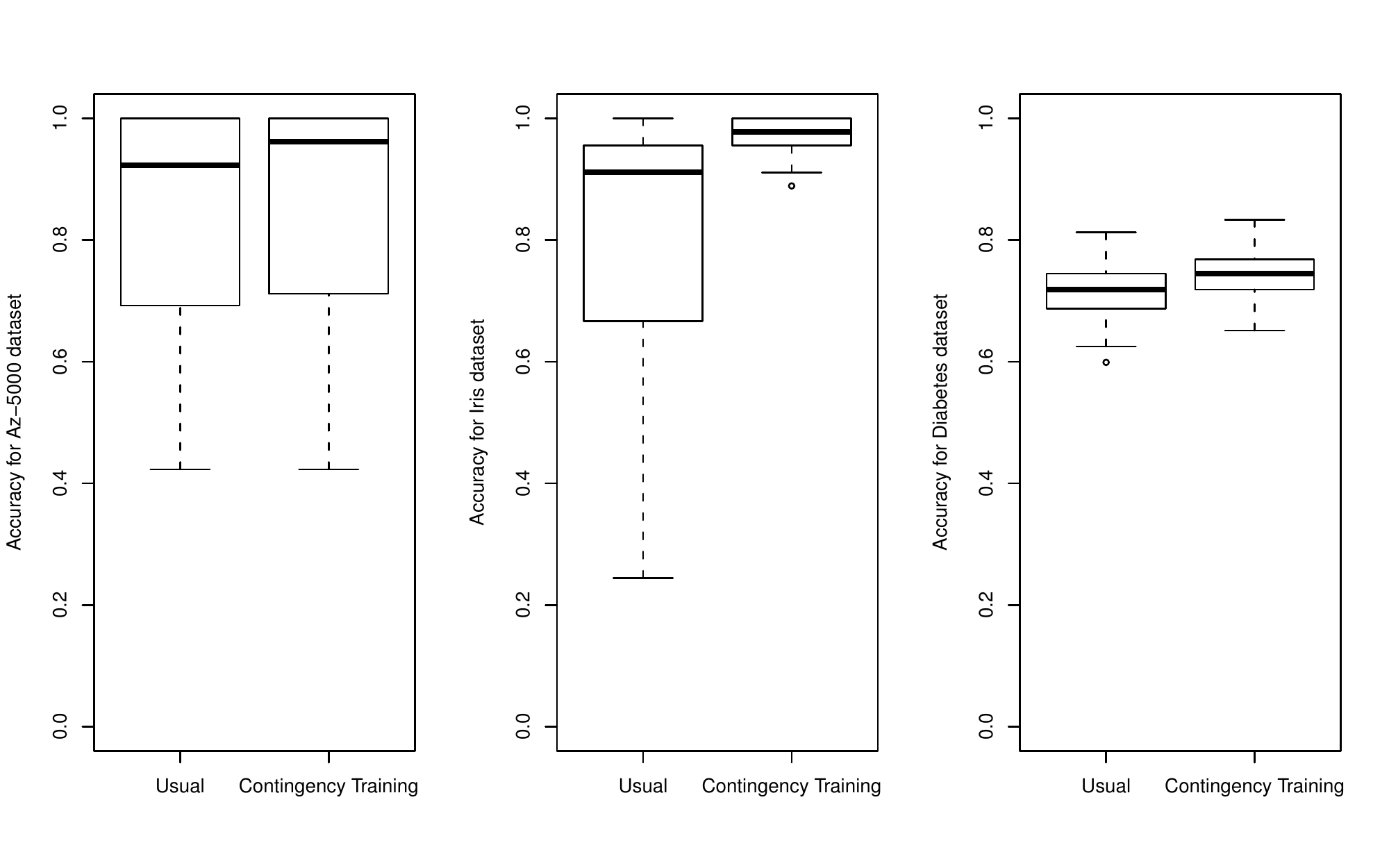}
 \caption{Accuracy on (from left to right) Zoo, Wine and Diabetes datasets using an unmodified dataset (usual training) and contingency training with a $10A/0I/0.1$ setting.}
 \label{challenge_10_0}
\end{figure*}

The objective of this section is to test the algorithm with datasets which has only a few or no irrelevant variables and check if the performance remains the same.
Figures~\ref{challenge_1_1} and~\ref{challenge_10_0} show the results for the usual training (training with the unmodified dataset) and contingency training for two different settings of the algorithm.
Contingency training performed similarly and even slightly surpassed the usual training for the majority of the tests.
Furthermore, the similarity of both Figures reveals that it is possible to replace the original dataset entirely for artificially created dataset (e.g. in the setting $10A/0I/0.1$) and still obtain equally improved results.

\section{Explanation}
\label{explanation}



The proposed method creates a mixture of constraints by removing information from the samples. 
An interesting feature of this mixture of constraints is that they possess zero bias. 
This happens because they are generated by the procedure $replaceWithMissing(S)$ which selects non missing values uniformly at random for the replacement with missing values.
This point diverges from the common use of constraints where they induce a bias over the learning procedure.

Furthermore, in the contingency training constructed dataset it may be even impossible to satisfy all constraints at the same time and the learning procedure must find a compromise that respect the majority of them. 
Finding this compromise is possible because the error is being measured over the sum of the errors of all the constraints and therefore the less constraints the classifier violates the better.
Thus, no constraint configuration is forced, but the algorithm itself will indirectly search for a best set of constraints.
Naturally, since the constraints created can not be entirely satisfied, the learning procedure must weight their importance and satisfy as much as possible.
For example, if some variables are less important than others the constraint of them being zero may be satisfied more easily.

To exemplify, consider the simple example of a system with two inputs and one output. 
The samples would be of the form:
\begin{equation}
	\{x_1, x_2, y\}
\end{equation}
With a small probability of substituting the values of the inputs with zero, the following possible samples can be artificially built:
\begin{align}
&\{x_1, x_2, y\} \\
 &\{0, x_2, y\} \\
 &\{x_1, 0, y\} \\
 &\{0, 0, y\}
\end{align}
The probability of the samples are from top to bottom $1 -2prob -prob^2$, $prob$, $prob$, $prob^2$.

Lets consider two situations to highlight how the method works.
First case, suppose both variables are important and necessary to predict $y$. 
Then, all artificially created samples would remain with a big prediction error.
Second case, suppose variable $x_1$ is not important to predict $y$.
The last two artificial samples will remain with a big prediction error.
However, both the first and second samples can be predicted accurately.
In other words, in this case both the first and second samples can be satisfied at the same time.

\section{Conclusions}

This article proposed a method for training classifiers called contingency training.
Contingency training works by removing information from the initial dataset for the creation of a bigger and less informative artificial dataset.

For the analysis of the proposed method, tests were conducted over traditional datasets, datasets with irrelevant variables.
Contingency training presented superior performance on datasets with irrelevant variables and similar accuracy on remaining datasets.

Moreover, from the tests it was delineated some explanation of how the proposed method achieves its performance. 
In summary, contingency training was shown to possess two nice direct or indirect implications:
\begin{enumerate}
\item Feature importance weighting by the satisfaction of constraints;
\item Absence of bias.
\end{enumerate}
To achieve this, the method only adds a small computational overhead caused by an augmented dataset.


Thus, with its accuracy improvement, ease of implementation and the capability of being applied to any classifier, contingency training is a promising algorithm that should soon be employed in applications in the real world.
Nonetheless, tests with other classifiers, different missing value representations and use in an ensemble of classifiers remain as future work.

\bibliographystyle{ieeetr}
\bibliography{genetic_algorithm}



\end{document}